# Autonomous Drone Swarm Navigation and Multi-target Tracking in 3D Environments with Dynamic Obstacles

**Suleman Qamar[1], Saddam Hussain Khan[1,2], Muhammad Arif Arshad[1], Maryam Qamar [3], Asifullah Khan*[1,2,4]**
[1]Pattern Recognition Lab, Department of Computer Information Sciences, Pakistan Institute of Engineering Applied Sciences, Nilore, Islamabad 45650, Pakistan
[2]PIEAS Artificial Intelligence Center (PAIC), Pakistan Institute of Engineering Applied Sciences, Nilore, Islamabad 45650, Pakistan
[3]Department of Computer Science & IT, University of Azad Jammu and Kashmir Muzaffarabad, Pakistan
[4]Center for Mathematical Sciences, Pakistan Institute of Engineering Applied Sciences, Nilore, Islamabad 45650, Pakistan

Corresponding author: Asifullah Khan (e-mail: asif@pieas.edu.pk ).

**ABSTRACT** Autonomous modeling of artificial swarms is necessary because manual creation is a time-intensive and complicated procedure which makes it impractical. An autonomous approach employing deep reinforcement learning, is presented in this study for swarm navigation.In this approach, complex 3D environments with static and dynamic obstacles along with resistive forces (like linear drag, angular drag and gravity) are modelled for tracking multiple dynamic targets. Moreover, reward functions for robust swarm formation and target tracking are devised for learning complex swarm behaviors. Since the number of agents is not fixed and have only the partial observance of the environment, therefore, swarm formation and navigation becomes challenging task. In this regard, the proposed strategy consists of three main phases to tackle the aforementioned challenges: 1) A methodology for dynamic swarm management, 2) Avoiding obstacles, Finding the shortest path towards the targets, 3) Tracking the targets, and Island modelling. The dynamic swarm management phase translates basic sensory input to high-level commands to enhance swarm navigation and decentralized setup while maintaining the swarm's size fluctuations. While, in the island modeling, the swarm can split into individual sub-swarms according to number of targets, and conversely, these sub-swarms may join to form a single huge swarm, giving the swarm ability to track multiple targets. Customized state-of-the-art policy-based deep reinforcement learning algorithms are employed to achieve significant results. The promising results show that our proposed strategy enhances swarm navigation and has the ability to track multiple static and dynamic targets in complex dynamic environments.

## I. INTRODUCTION

Deep reinforcement learning exploits the ideas of deep learning [1, 2], and reinforcement learning [3]. It has been used for learning advantageous behaviors for agent training [4]. The technique of integrating simple activities in such a manner that a more sophisticated behavior develops is known as swarm intelligence [5, 6]. It has allowed us to replicate numerous natural processes and relatively simple species cooperating and completing complex tasks to achieve excellent results in a variety of disciplines while ostensibly conducting simple activities. As a result of its utility, engineering artificial multi-agent systems with swarm intelligence has become a burgeoning field of study. Swarm intelligence has a wide range of uses [7, 8], including high-level monitoring of dynamic networks, adaptive routing in telecommunications, distributed sensing technology [9], surveillance [10], data processing, cluster analysis, search and rescue missions [11, 12],advertisement, and drone used as a delivery bot [13]. This is especially important in current times of pandemic and there are also potential applications such as using nanobots within the body of a cancer patient to kill tumors [14]. Swarm intelligence is also being employed by NASA in planetary imaging [15]. Swarm intelligence is a natural phenomenon in which the activities of numerous dispersed and self-organized simple organisms combine to produce "intelligent" global behavior. Swarm awareness may be observed in nature in several subtle and awe-inspiring ways, such as when simple species acting independently interact to generate complex global behavior. Bee colonies, schools of fish, flocks of birds, ant colonies, hawks hunting, animal herding, and bacterial growth are all examples of swarm intelligence [16]. To fulfill common objectives like foraging, group coercion, and alignment control, these swarm systems employ the notions of "quantity" and "coordination" [17, 18]. Swarm robotics [19, 20] is a technique for coordinating many robots as part of a larger structure made up mostly of basic physical robots [21]. It is expected that robot-robot interactions and their interactions with the environment result in intended reciprocal behavior [22]. Artificial swarm intelligence, as well as biological observations of flies, ants, and other natural systems that display swarm behavior, inspired this work. However, most artificial swarm systems find it challenging to represent such a mix of behaviors displayed by natural swarms, since doing so adds to the problem's complexity [23, 24]. When the drones are controlled manually, an operator must be present to operate it. Furthermore, traditional machine learning requires manual feature engineering which is tedious and less flexible. This research work addresses the problem of the development of an end-to-end model for detecting targets in various settings and autonomous navigation of drones swarming towards them while avoiding obstacles and maintaining stable agent formations.

To the best of our knowledge, research conducted in this field lack the following:
1) Finding optimal paths for multi-swarm navigation and obstacle avoidance in unseen complicated 3D environments.
2) Dividing a swarm into numerous sub-swarms and combining these sub-swarms back to form a large single swarm based on the number of targets and distance between them.

Overall, this work has the following contributions:
1) A policy-based deep reinforcement learning strategy is proposed which enables the drone swarm to navigate autonomously while avoiding obstacles. To prepare the drone swarm for real-life situations, complex 3D environments with dynamic obstacles having distinct morphology are created. In addition, resistive forces like linear drag, angular drag, and gravity are added to make the environments more realistic and complicated.
2) To improve swarm navigation and decentralised setup while preserving the swarm's size variations, a mechanism that converts basic sensory input to high-level commands is employed. The concept of memory is also added to aid drone swarms in remembering best paths.
3) Novel reward functions have been introduced that allow the swarm to avoid barriers and identify targets while traversing the shortest path and maintaining a stable swarm structure. Both static and mobile targets can be efficiently tracked.
4) Multi-target tracking is introduced, where our swarm can track multiple targets and still maintain communication. Multi-swarms are also introduced, where a swarm can divide into sub-swarms to track multiple distant targets simultaneously, and when they come close, sub-swarms can combine again while maintaining stable connections with each other, this is named as Island modelling.

This research aims to create artificial swarms for the purpose of navigation in unseen environments and tracking of targets. Five key swarm behaviors are modelled: (1) Swarm formation and organization, (2) Dynamic obstacle avoidance, (3) Locating the target, (4) Navigation towards the target using shortest path while sustaining swarm formation, and (5) tracking multiple targets by dividing the swarm into subswarms and tracking each target with a single sub-swarm. The research is particularly focused on finding multiple targets in complex environments resembling real-world scenarios by training swarm agents using Unity 3D. The framework is shown in Fig.1.

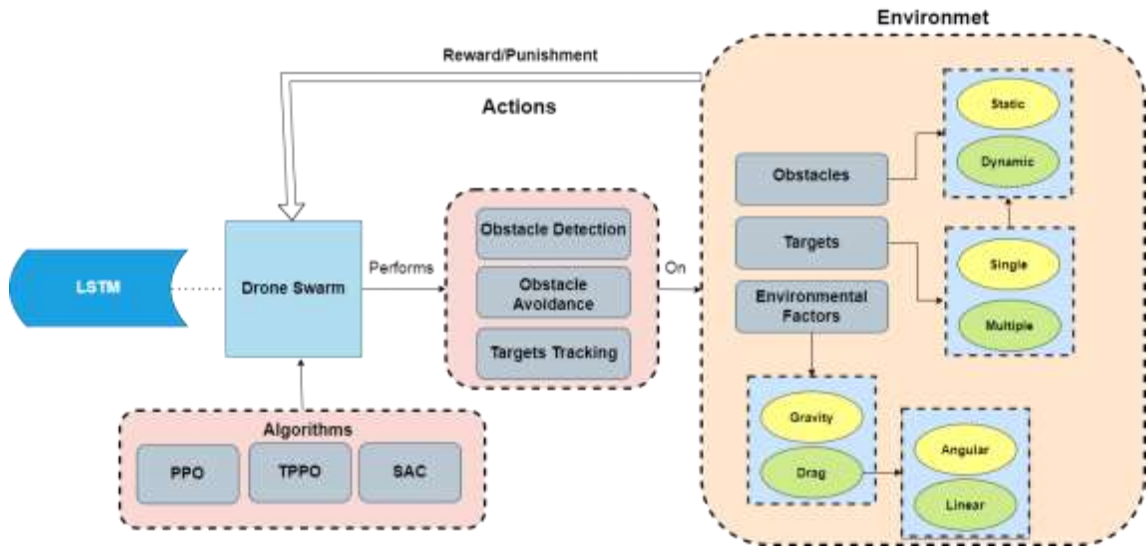

*1: Framework representing all components, their relationship and arrangement*

## II. RELATED WORK

Researchers have long been interested in extracting and implementing the principles that regulate these amazing biological swarm systems since their performance regularly surpasses individual biological organisms'. Swarm simulations and manual inspections have previously achieved substantial results [25-28]. Mimicking the swarm behavior of animals manually has been studied extensively, for example, in [29] the laws that control enormous prey recovery in insects were employed to explain the realization of swarm behavior in robots. Minaeian et al. [30] developed SLAM algorithms for a set of unmanned ground and aerial vehicles to map and monitor crowds. Pheromone-based localization of dispersed targets by a swarm of virtual agents operating in a simulated discretized environment was studied in [31]. In their research, mini-UAVs are viewed as swarm agents, and they may have imperfections while detecting targets. Swarm behaviors, such as aggregation, foraging, creation, and monitoring, were studied in [32] and algorithms were developed to replicate such behavior.

When the problem's complexity grows exponentially, the time and effort required to solve it also grows dramatically.So, the time and effort spent inspecting, formulating, and solving a problem must be reduced. Q-Learning was developed by Watkins et al. [33] in 1992 as a strategy for iteratively training agents to behave optimally to maximize reward. With sufficient samples, Q-Learning converges to produce optimum action-value pairings in Markovian domains with a probability of one. Google Deep Mind built the first deep learning model based on Q-Learning (DQN) in 2013 [34], it could play a variety of Atari 2600 games using only pixels as feedback. However, in the Q-Learning algorithm, state space is continuous, but action space is discrete, so it can't be used in problem domains where action space also needs to be continuous.

To handle the problem of continuous action space, Lillicrap et al. [35] proposed a deep deterministic policy gradient-based actor-critic algorithm (DDPG) that borrows heavily from DQN in terms of simple architecture, including mini-batch updates and the Ornstein-Uhlenbeck process [36] as exploratory noise. Swarm formation and mutual localization was explored in [37, 38], with a few modifications, they utilized the DDPG algorithm. Actor-network and critic-network input was adapted to make the swarm machine work. A novel technique was utilized in which they gave the critic network entire state data while only giving the actor network partial state data. As a result, the critic network uses global state data to modify the parameters of the actor network, whereas the actor network simply uses local state data.

In the technique proposed by Akhloufi et al., a deep learning method is provided to anticipate the behavior of agents tracking a travelling drone [39]. A single agent was trained by [40] to maneuver in a dynamic maze-like environment using deep reinforcement learning. In their research, information from location sensing devices was employed as feedback to train the model with memory.

An iterative technique called Trust Region Policy Optimization (TRPO) [41] was introduced that operated similarly to natural policy gradient methods and could optimize large non-linear policies. Thus, it could be effectively used for neural networks but required second-order derivative calculations. Proximal Policy Optimization (PPO) [42] from Google Deep Mind became a state-of-the-art solution to train agents because of its sample reliability. It largely followed the concept behind TRPO but reduced calculations from second-order derivative to the first-order derivative. It used a stochastic gradient ascent to optimize a "surrogate" clipped objective function. In the Atari domain, the PPO algorithm out-performed Advantage Actor Critic (A2C) [43]and Actor Critic with Experience Replay (ACER) [44]. PPO algorithms were used to measure the efficiency of OpenAI Gym on high-dimensional controls, such as humanoid running and steering. In [45] code level optimizations for the PPO algorithm to work properly were summarized. A variant of the PPO algorithm named IEM-PPO [46] was presented with improved sample efficiency, better stability, and robustness, yielding comparatively higher cumulative reward, but took more time to train. PPO algorithm along with incremental curriculum learning [47] and long-short-term memory (LSTM) [48][55] was utilized to implement an adaptable navigation algorithm. The Truly Proximal Policy Optimization (TPPO) [49] modifies the PPO algorithm to perform slightly better in terms of stability and sample efficiency. Hämäläinen et al. [50] argued that in PPO, the variance of exploration prematurely shrinks, which makes progress slower, and proposes PPO-CMA to dynamically increase or decrease the variance of exploration. A new surrogate learning objective featuring an adaptive clipping mechanism named as PPO-$\lambda$ is introduced in [51]. It iteratively improves policies based on a theoretical goal for adaptive policy change. PPO algorithm was employed in [52] to create drone swarms using multiple sensors per agent to reach target information while avoiding obstacles. Wang et al. [53] worked on an autonomous multi-agent target reinforcement model using UAVs for searching and tracking. Camera, IMU and GPS were used as sensors. Since their work was based on patrol, they only considered a constrained environment. Qu et al. [54] leveraged the concept of association for the purpose of multiple target tracking. Their research centered on intelligent sensors that can distinguish readings based on targets by using Grideye, an infrared sensor with ability to calculate target location and surface temperature. Multi-target tracking on MOT15 and MOT16 datasets is performed by Ren et al. [55].

PPO despite having good performance suffered from sample efficiency. It showed good performance On small sample spaces, however low sample efficiency caused a lot of problems on large sample spaces. TPPO enhanced sample efficiency but improvements were still limited. To counter this problem, an off-policy method known as Soft Actor-Critic (SAC) was introduced in [56] that focused on maximizing reward with maximum possible entropy which is the measure of randomness. SAC required extensive hyperparameter tuning in some cases, but achieved state-of-the-art results.

## III. METHODOLOGY

Due to the complexity of the current world and the comparatively poor sample efficiency of algorithms in the field of deep reinforcement learning, it is difficult to directly create a real-life model. Furthermore, explicitly training our model in the real world might result in mishaps. Thus, models are trained using simulations, utilizing Unity3D engine since it provides the required tools, needed to construct complex 3D environments. Another reason is that it includes the ml-agents library, which allows Python to be used as a backend for deep learning tasks. Architecture is represented in Fig. 2.

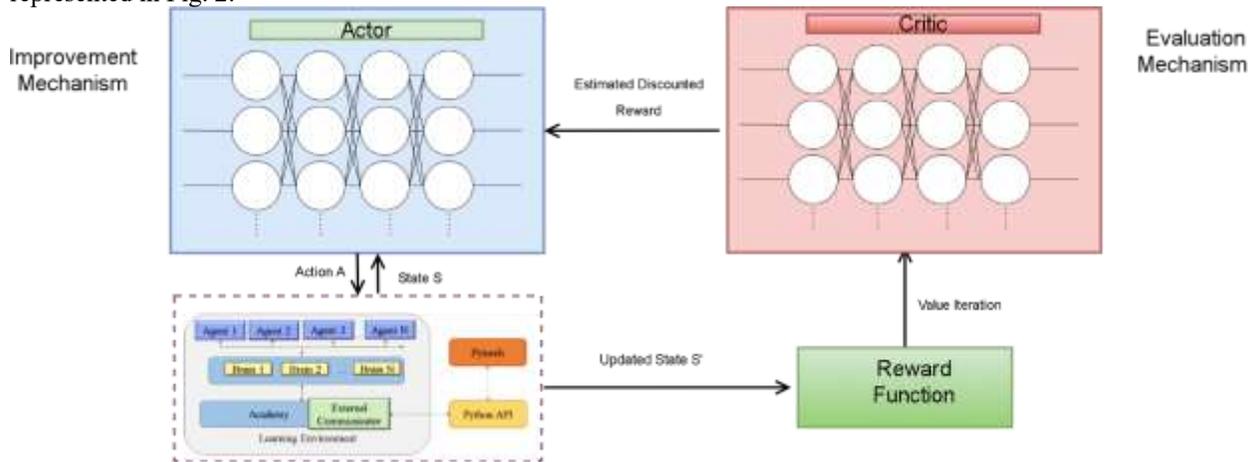

2: Actor-Critic along with Simulation Environment

### A. SIMULATION ENVIRONMENT

It's crucial to design environments that support agents' ability to train successfully and efficiently. Therefore, a variety of scenarios are constructed for training the drone swarm to determine the best conditions. Agents with their sensing area and different levels of danger identification by agents is visualized in Fig. 3.

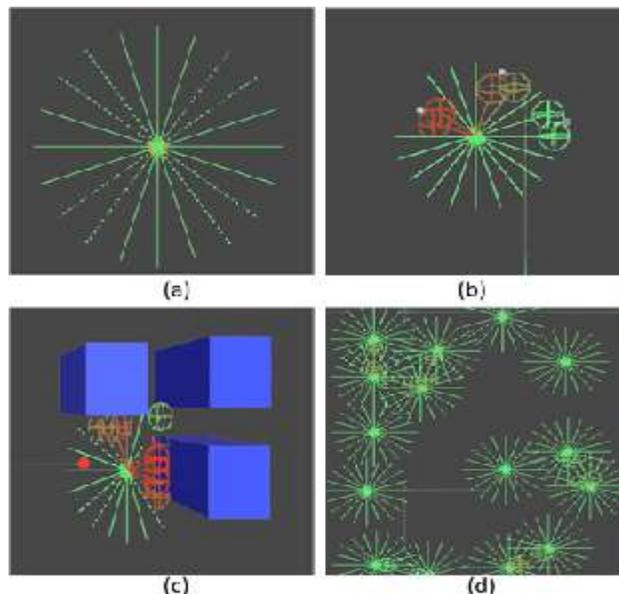

3: (a) Agent with its sensing area (b) Three levels of danger identified by agent, "Green" represents slight danger; no immediate action required, "Brown" represents intermediate danger; try to minimize it without sacrificing goal; "Red" represents extreme danger; prioritize safety (c) Agent's interaction with obstacles (d) all agents maintaining safe distance with each other

Basic environments are visualized in Fig. 4(a) whereas agents within environment is given in Fig. 4(b). Several settings are used during training to extend our model and assess the efficacy of different training circumstances. All models include a 3D environment with a variable number of obstacles that cover various cubical volumes as indicated in Table 1.

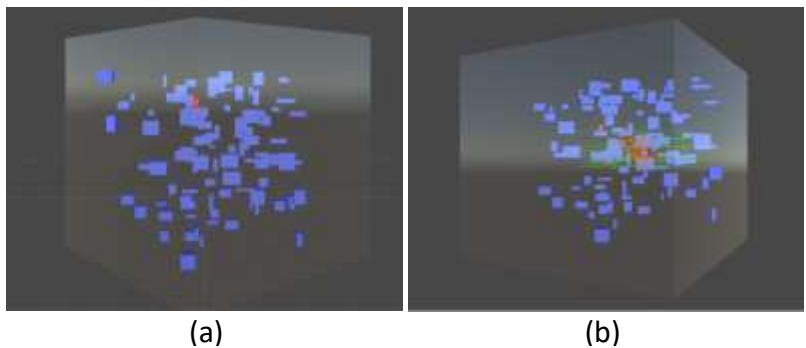

(a)  (b)

4: Environment visualization: grey box represents outer boundary (a) obstacles are blue colored while target is represented by a red colored capsule (b) Agents with their obstacle lines (green in color) are visualized

Obstacle locations are chosen at random to make our method adaptable to changing surroundings. The number of multi-agent training targets varies across simulations, ranging from two to sixteen. Table 2 lists the environments utilized for multi-target training. Summary of actions that can be taken by agents are listed in Table 3.

Table 1: Single target environments

| Volume in units $^3$ | Length of each axis in units | Number of obstacles | Length of obstacles in units | Obstacle positions |
|---|---|---|---|---|
| 1,000,000 | 100 | 100 | Ranging between 1 to 10 | Random |
| 500,000 | 50 | 60 | Ranging between 1 to 5 | Random |
| 100,000 | 10 | 30 | Ranging between 1 to 7 | Random |

Table 2: Multi-target environments

| Volume in units $^3$ | Axis Length | No. of obstacles | Obstacles' Sizes | Obstacle positions | No of Targets | Target Type |
|---|---|---|---|---|---|---|
| 1,000,000 | 100 | 20 | Ranging between 1 to 10 | Static | 2 | Static |
| 500,000 | 50 | 50 | Ranging between 1 to 5 | Random | 2 | Dynamic |
| 100,000 | 10 | 30 | Ranging between 1 to 7 | Random | 4 | Static |
| 100,000 | 10 | 10 | Ranging between 2 to 8 | Static | 8 | Dynamic |
| 700,000 | 70 | 100 | Ranging between 5 to 10 | Random | 16 | Static |
| 1000,000 | 100 | 200 | Ranging between 1 to 10 | Random | 16 | Dynamic |

The drawback of using only static targets throughout the training period caused the model to simply memorize their locations resulting in loss of generalization. To counter this, the position of targets in the environment is also randomized by employing the random tick, which involves changing the positions of target points every 100 ticks to introduce time-decoupled uncertainty. To generate ticks, a random floating-point value is generated at each time-step. The threshold for ticks being considered is 0.85.

*Table 3: Allowed Actions*

| S.No. | Action | Possibilities |
|---|---|---|
| 01 | Motion | 9 Directions (up, down, right, left, forward, backward, pitch, roll, yaw) |
| 02 | Rotation | [0-360) |
| 03 | Hover over target | Yes/No |

Agents must also begin each simulation in a random place, which necessitated the creation of a position randomizer. It generates a random position in the environment from all available places. Obstacles, targets, and drone agents that have already been created are all eliminated from prospective locations. Each agent utilizes distance sensors in a circular pattern around them to detect and avoid obstacles. Our model is faster and uses fewer resources, because these sensors eliminate the need for significant processing, which is common in camera-based techniques.

Every agent has knowledge about the targets and their immediate surroundings. At each time-step, the positions of the targets are assumed to be known. A Sigmoid of the normalized geodesic distance between the targets' and the agents' positions is taken to generate a compact representation of the targets' locations. For ease of use, all agents and targets have an aligned coordinate system. Also, their rotational characteristics are locked. The following formulae are used to calculate the relative orientation standardized vector **P** as shown in Eq.1. The distance between target locations **i** and agent y are then computed by taking their sigmoid as shown in Eq. 2.

$$\mathbf{P} = \left(\frac{S_i - S_y}{||S_i - S_y||}\right) \quad (1)$$

$$\sigma_{b_i}^{y} = \left(\frac{||S_i - S_y||}{1 + ||S_i - S_y||}\right) \quad (2)$$

where $S_i$ and $S_y$ denote the target points and agent y's location, respectively. The location information related to all agents present in the nearby zone and obstacle sensor outputs are included in the information state vector. As agent numbers in a neighborhood might fluctuate and deep neural networks (DNNs) have finite input capacity, feeding location data from nearby agents directly into the network can be difficult. To address this problem, a 3D histogram approach is proposed. In the first step, vectors are computed by subtracting the position of agent y from all agents surrounding it, and then these vectors are mapped into J bins in each axis. Then the axes are concatenated into a single vector and given to DNN as input. Vector component **<k>** histogram computation for every agent is calculated using an algorithm called HVC presented as follows:

Algorithm 1: HVC
$His_j^{<K>} \leftarrow 0, j = 1,2,3,\ldots J$ *(Initialize)*
Input:
x: number of agents; $D_c$: Communication Radius
Output:
Vector Component Histograms

For x=1 to W do

$$j = \left[\frac{<K>_y^x + D_c}{D_c \times 2}\right], g \in \{x, y, z\}$$

$D_s \leq b_y^x \leq D_c$ $\qquad His_j^{<K>} = \frac{1 - \sigma_y^x + D_c}{W \times 3}$

Otherwise $\qquad\qquad\qquad 0$

**x** denotes all agents except agent y, the difference in **k** components of position vectors of agents **x** and y is referred to as $<k>_y^x$. Distance between agent x and y is denoted by $b_y^x$; $D_s$ is safe distance parameter while $D_c$ represents a communication area. Obstacle detection sensors denoted by $(D_{sen})$ are considered time-of-flight $(ToF)$ sensors. Output of all sensors is **Ti** represented by Eq 3.

$$T_i = \begin{cases} \frac{1-c_i}{D_{sen}}, & \text{if if an obstacle is sensed by "i"} \\ 0, & \text{otherwise} \end{cases} \quad (3)$$

After that, the data from all the J sensors are aggregated and submitted to the DNN.

### B. ACTION GENERATION

A delta vector **u** composed of three variables, representing 3D coordinates, is generated. The vector **u** and the agent's present position vector $S_i$ is then summed to get new position $S_i$ as shown in Eq 4.

$$(S'_i) = S_i + U \quad (4)$$

This action generation strategy broadens the applications of the proposed approach.

### C. PERFORMANCE METRICS

The performance metrics for assessing successful training of the model are explained in this section. **Mean Cumulative Reward (MCR)** is an assessment of total reward which indicates an increasing trend during effective training.
**Value Loss (VL)** is the performance indicator for assessing policy change, registers a declining trend during effective training.
**Policy Loss (PL)** is an accurate state value prediction which exhibits growing trend until the reward is stable, then it starts to drop during effective training. **Entropy (E)** represents the unpredictability of the model's decisions and exhibits a diminishing trend during effective training.

### D. MODELLING REWARDS

Training data is acquired by running multiple simulations in parallel and then optimizing our novel reward function using PPO, TPPO and SAC. $R_n$ (5) is the navigation reward that allows a swarm of drones to calculate the shortest path between the target and the swarm in real-time, even if the target is behind a large obstacle with no direct path to it. Shortest path is determined at each timestep, while the target is moving and drones are in flight, so that drones may surround the target as quickly as possible. $R_o$ (6) denotes the reward for assisting in the construction and organization of drone agents. It enables them to form swarms in real-time while remaining at a safe distance yet close enough to communicate. With the assistance of $R_s$, the swarm can avoid obstacles (7). In addition, if an agent leaves the environment or if the agent is destroyed, then a negative reward as punishment is generated, and the destroyed agent is respawned at some random place inside the environment. Total reward by a single swarm is represented using RS (8) which is a combination of individual rewards like $R_n, R_o, R_s$. $R_{ms}$ (9) rewards all swarms that are present in the environment and are cooperating. Swarm divides into sub-swarms and similarly multiple swarms combine to form a single swarm. The number of targets determines how the swarm subdivision is done. If there is one target with a swarm tracking it, and another target is introduced, the swarm will split to track both targets. Similarly, increasing the number of targets increases the number of swarm sub-divisions. Also, if two swarms are tracking two targets and one of the targets is eliminated, the two swarms merge to form a single swarm.

$$R_n = \begin{cases} 1 - \sigma(b_y^i - D_s), & if\ b_y^i \in [D_s, \infty) \\ 1, & \text{otherwise} \end{cases} \quad (5)$$

$$R_o = \begin{cases} \frac{1-\sigma(b_y^i - D_s)}{M \times 3}, & if\ b_y^x \in [D_s, D_c) \\ -1, & \text{otherwise} \end{cases} \quad (6)$$

$$R_s = 1 - \sigma\left(\sum_{p=1}^{P} U_p\right) \quad (7)$$

$$RS = R_n + R_o + R_s \quad (8)$$

$$R_{ms} = \frac{\sum_{n=1}^{N} RS_n}{N} \quad (9)$$

### E. HYPERPARAMETERS

The following tables summarize the hyperparameters used during training and evaluation phases:

*Table 4: Simulation Hyperparameters (PPO, TPPO and SAC)*

| S. No. | Hyperparameter | Value |
|---|---|---|
| 01 | Simulation Instances | 28 |
| 02 | Agent Quantity in Simulations (S) | 23 |
| 03 | Number of steps per episode | 900 steps |
| 04 | Radius for Communicating ($D_c$) | 9 units |
| 05 | Histogram Bins per Axis (K) | 32 |
| 06 | Obstacle Sensors per Agent (J) | 18 |
| 07 | Sensor Range ($D_sen$) | 7 units |
| 08 | Safe Region ($D_s$) | 3 units |

*Table 5: Training Hyperparameters (PPO and TPPO)*

| S.No. | Hyperparameter | Value |
|---|---|---|
| 01 | Total steps | 55 millions |
| 02 | Time Horizon | 512 steps |
| 03 | Size of Batch | 1024 |
| 04 | Buffer Size | 10240 steps |
| 05 | Rate of Learning | 0.0007 |
| 06 | Policy update penalty (beta) | 0.007 |
| 07 | Clipping Value | 0.3 |
| 08 | Lambda | 0.96 |
| 09 | Epochs | 2 |
| 10 | Learning Rate Decay | Linear |

*Table 6: Training Hyperparameters (SAC)*

| S.No. | Hyperparameter | Value |
|---|---|---|
| 01 | Total steps | 55 millions |
| 02 | Time Horizon | 512 steps |
| 03 | Size of Batch | 256 |
| 04 | Buffer Size | 10240 steps |
| 05 | Rate of Learning | 0.0007 |
| 06 | Policy update penalty (beta) | 0.007 |
| 07 | Clipping Value | 0.3 |
| 08 | Lambda | 0.96 |
| 09 | Epochs | 2 |
| 10 | Buffer Initial Steps | 12 |
| 11 | Initial Entropy Coefficient | 0.9 |
| 12 | Save Replay Buffer | True |
| 13 | Tau | 0.005 |
| 14 | Steps per Update | 3 |
| 15 | Reward Signal Number Update | 3 |

The policy and critic networks, each with two dense layers of 64 neurons, are employed by the PPO, TPPO and SAC with learning hyperparameters given in Table 5 and Table 6. Architectures are optimized to facilitate faster convergence at about 10 million steps. Moreover, only two epochs were used to further cut down on training time.

### IV. RESULTS AND DISCUSSION

Performance curves for PPO, TPPO and SAC are obtained for 55 million training steps.

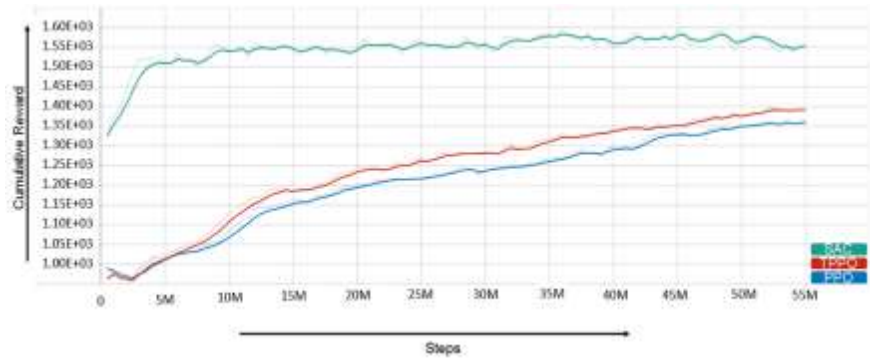

(a)

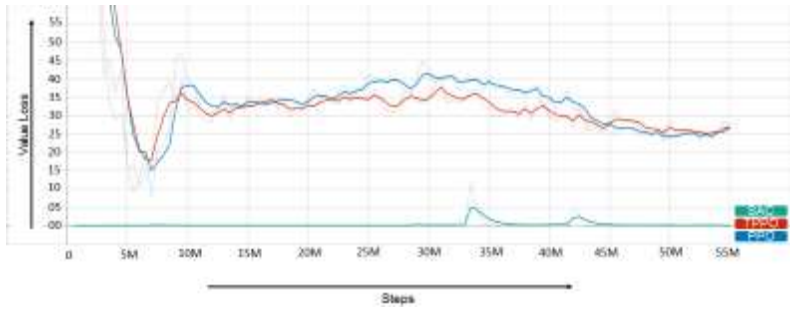

(b)

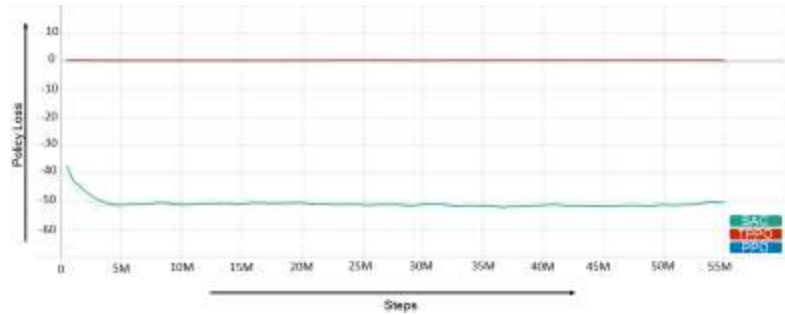

(c)

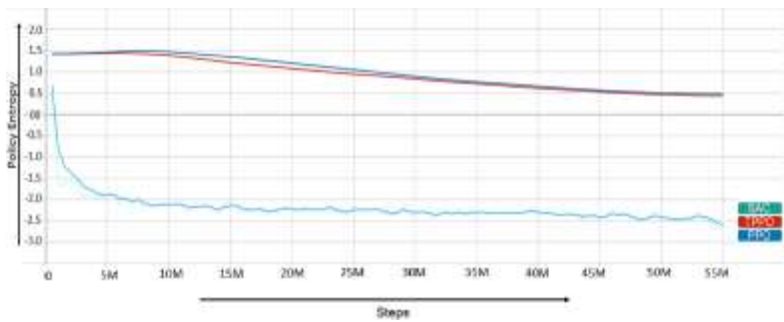

(d)

*5: Comparison of SAC, PPO and TPPO in terms of (a) Mean cumulative reward (MCR) (b) The value loss (y-axis) against 55 million training steps (x-axis) (c) Policy Loss*

Comparative Cumulative Reward (CR) is provided in Fig. 5(a). The increasing trend shows successful training. SAC performs best due to it being sample efficient, while PPO and TPPO have comparable results. Policy Loss (PL) and Value Loss (VL) curves are given in Fig. 5(b) and Fig.5(c), the performance of SAC can be visually verified. In Fig. 5(d), Entropy, change in policy during training, is shown. Entropy shows decreasing trend which correlates to successful learning.

A visual assessment of the swarm's activity is required to demonstrate the feasibility of our strategy. The visual depictions of several aspects of a swarm's activity are given in experiment section (Demonstrations can be accessed here[1]). Spheres indicate the agents, while target is represented by black and sometimes red color. Everything else is included in obstacles. When two or more agents are in each other's communication zone, a green-colored edge forms between them to visually demonstrate communication. The drone agents not only effectively form formations in the demos, but they are also able to navigate the complicated terrain, avoiding different sized and shaped obstacles while maintaining a swarm-like structure.

Furthermore, it successfully reaches and surrounds the targets. Even if the target is constantly moving, the swarm continues to encircle it. When several dynamic targets are traveling in various directions, the swarm splits into sub-swarms, with each sub-swarm having the same number of agents following a single target object. Sub-swarms merge to form a bigger swarm when targets approach closer to one another.

## V. EXPERIMENTS AND ANALYSIS

Experiments are carried out to test and develop our models. The complexity of the environment is gradually raised, and model improvements are made through gaining insights from the model's undesirable actions. Experiment summary is given in Table 7.

### A. EXPERIMENT 1: SWARM ORGANIZATION AND OBSTACLE AVOIDANCE

For evaluation of swarm formation and observe swarm-like behavior, a 1000 units$^3$ obstacle-free environment is created as shown in Fig. 6(a). Drones successfully formed a swarm and surrounded the black target visualized in Fig 6(b). Moreover, target was made mobile while keeping all other factors constant, swarm again surrounded the moving target as shown in Fig. 6(d). Next, Three large green obstacles, a black target and 18 blue colored drones are placed in the environment and drones form a swarm and successfully converge surrounding the target visualized in Fig. 7(a) and 7(b), respectively. Also, the target was relocated behind an obstacle to evaluate obstacle avoidance mechanism of swarm as seen in Fig. 7(c). Swarm's ability to avoid obstacles while maintaining swarm formation, surrounding the target and tracking dynamic target can been observed in Fig. 7(d-h).

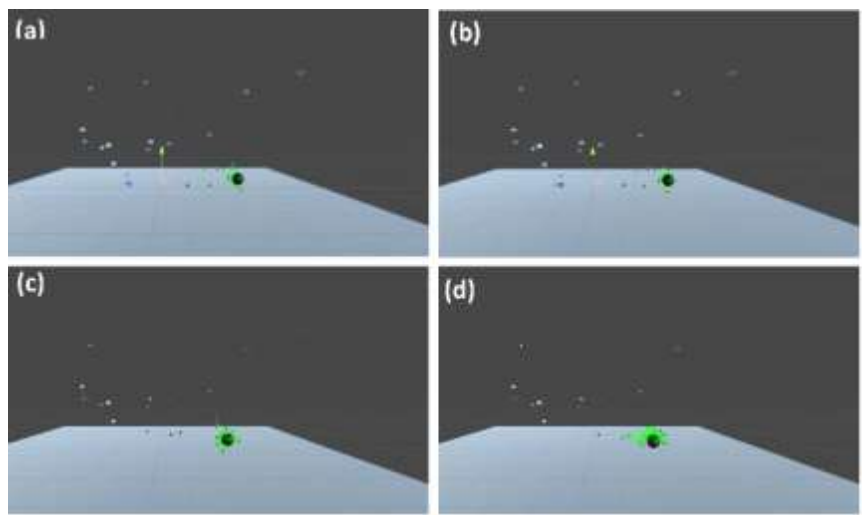

*6: Swarm formation, organisation, maintenance while surrounding the target and tracking the moving target, maintenance*

---

[1] https://www.youtube.com/playlist?list=PLq0872kWvR0VqTcnWDrudsDRELvt_FY7w

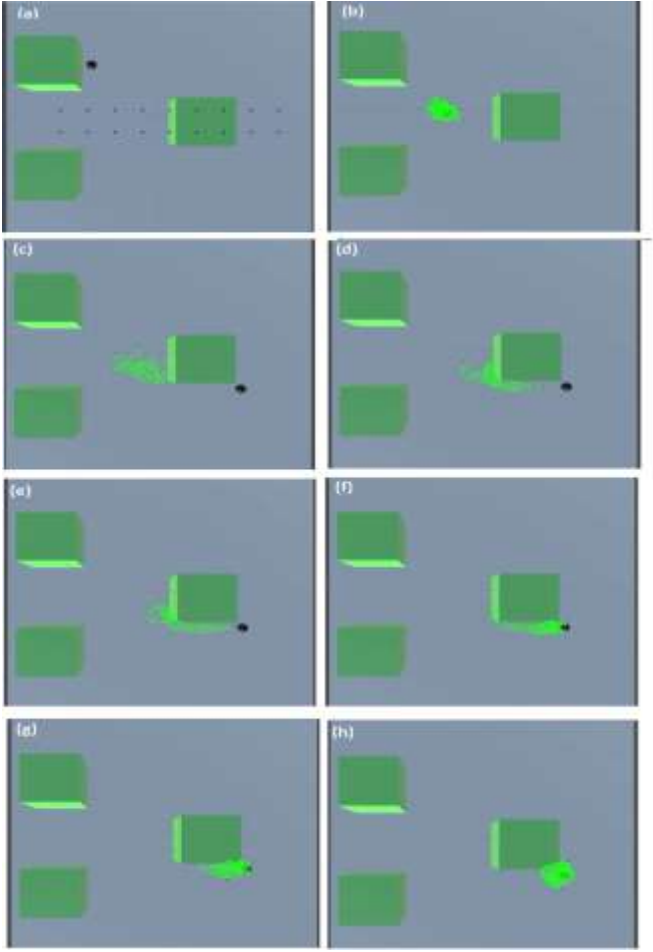
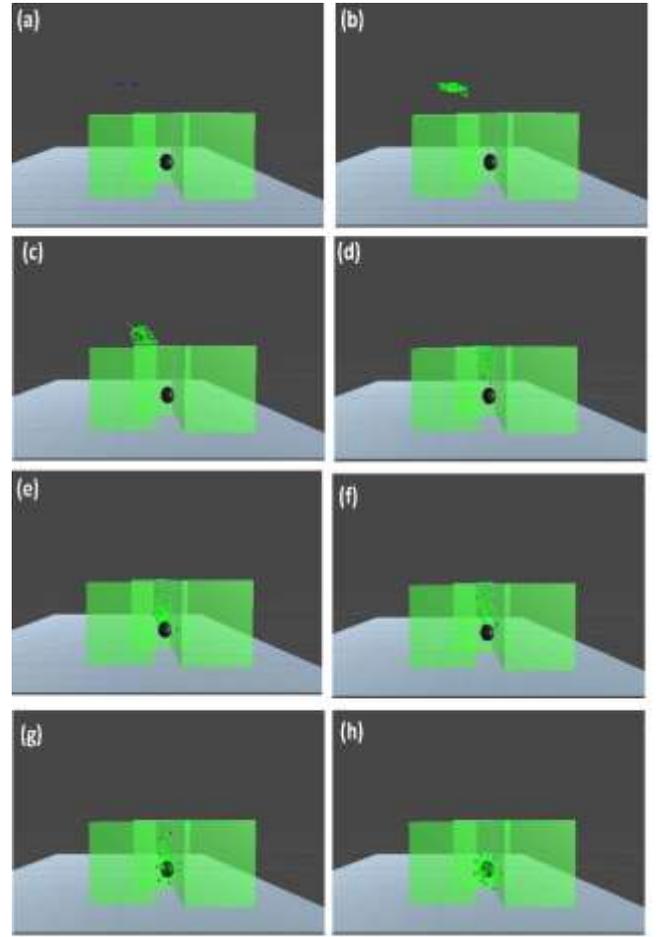

*7: Swarm Organization and Obstacle Avoidance*  *8: Target surrounded by obstacles*

### B. EXPERIMENT 2: TARGET SURROUNDED BY OBSTACLES

In this experiment, the target object was placed behind a series of obstacles (transparent green in color) and given a single convoluted path to transverse in order to reach the target shown in Fig.8(a-b). The swarm's trajectory demonstrates intelligent activity. The swarm body moves around obstacles trying to find an opening and when found, swarm quickly surrounds the target visualized in Fig. 8(c) and 8(d-h),respectively.

### C. EXPERIMENT 3: TARGET BEHIND AN OBSTACLE

In this experiment, drones' robustness is tested if they can find a target hidden completely behind an obstacle. The first experiment failed to provide ideal results since the swarm was unable to reach the destination as shown in Fig. 9. The issue was that the swarm was using Euclidean distance, which does not account for obstructions or alternative pathways. So, geodesic distance was introduced which calculates shortest distance along the manifold. Same experiment was re-performed with geodesic distance successfully (Fig.10). As seen in Fig. 10(c-f), swarm moves around the wall to reach the target while traversing shortest possible distance.

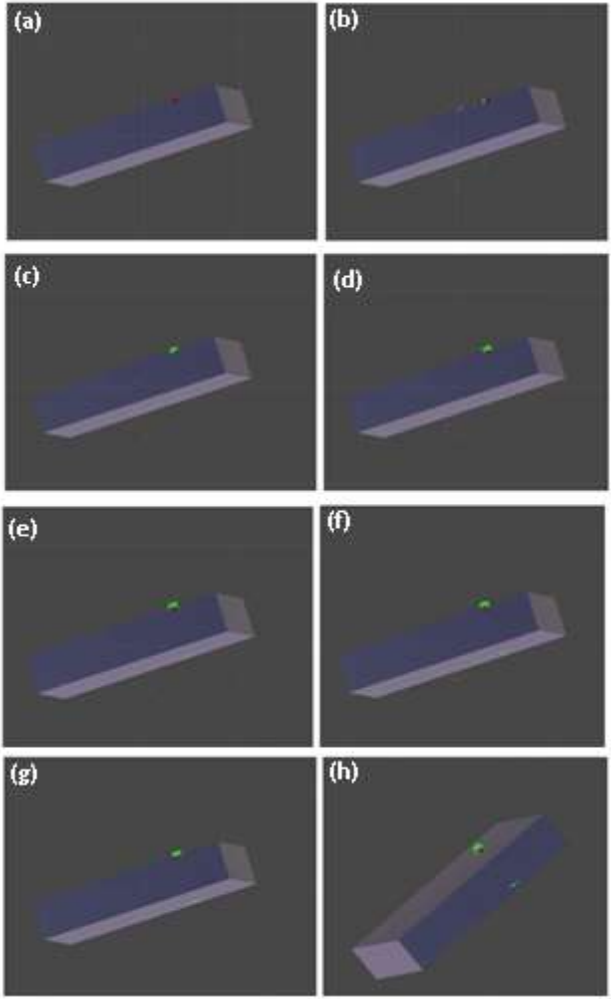
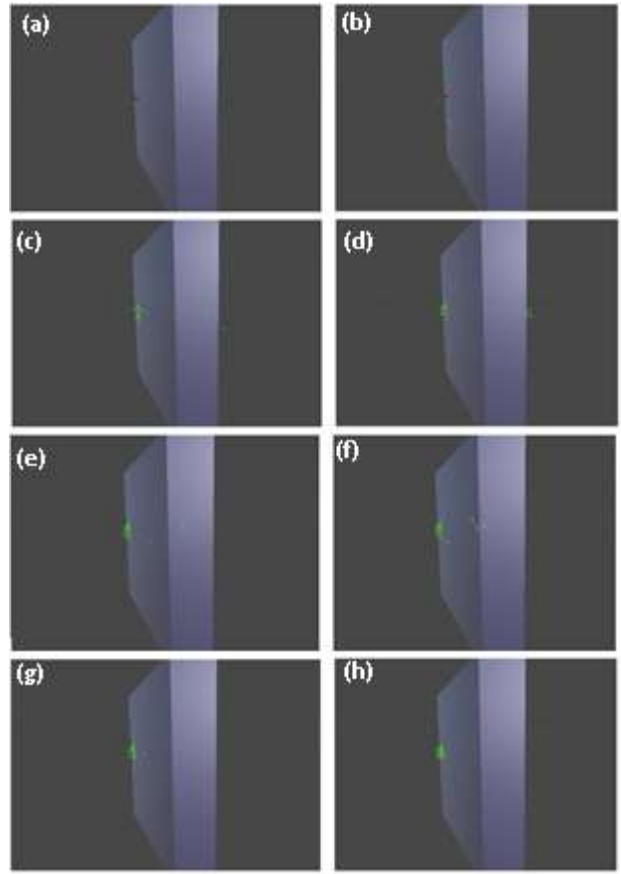

*10: Target behind an obstacle with no visible path*

*9: Target behind an obstacle with no visible path*

### D. EXPERIMENT 4: SWARM TRAVERSING THROUGH A HOLE IN SINGLE FILE

In this experiment, a hole was added in the wall which was present in the previous experiment. As the path through the whole has become the shortest path, the idea is to check if the model would make the swarm pass through the hole. To further increase the difficulty, hole was made small enough so that only one drone can pass through it at a time.

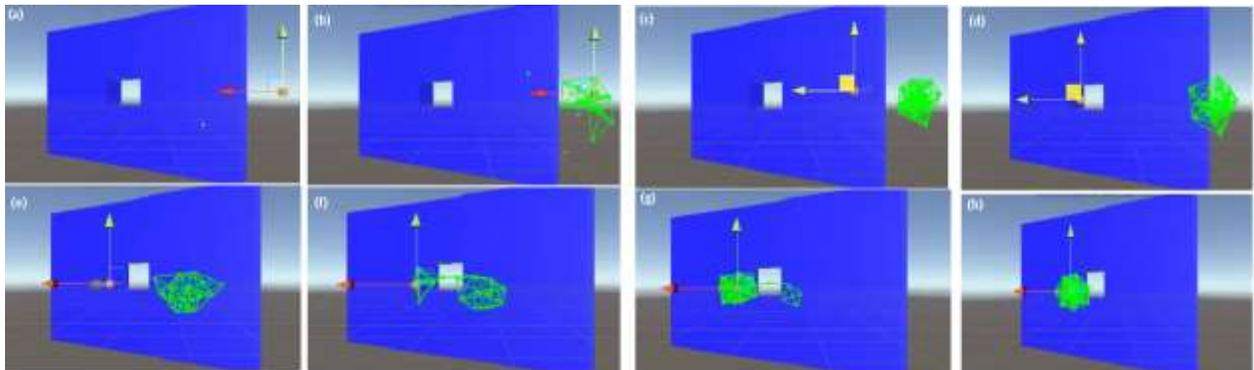

*11: Swarm traversing through a hole in single file*

Target was placed on one side of the wall (Fig. 11(a) and drones were allowed to converge around it and form a swarm (Fig. 11(b-c)). Then target was moved to other side of the hole (Fig. 11(d)). Swarm started to move through the hole, one drone at a time, and surrounded the target on the other side (Fig. 11(e-h)).

## E. EXPERIMENT 5: MULTIPLE SWARMS AND TARGET OBJECTS

In this experiment, two static targets were placed close to each other in the environment whose volume is 1000 units$^3$ (Fig. 12(a)). For visual reasons, the size of the targets was enhanced, but, for our swarm, they were simply dots in three-dimensional space in all experiments so 1 unit out of 1000,000,000 units. Swarm was successful in locating both targets and surrounding them (Fig. 12 (b-f)).

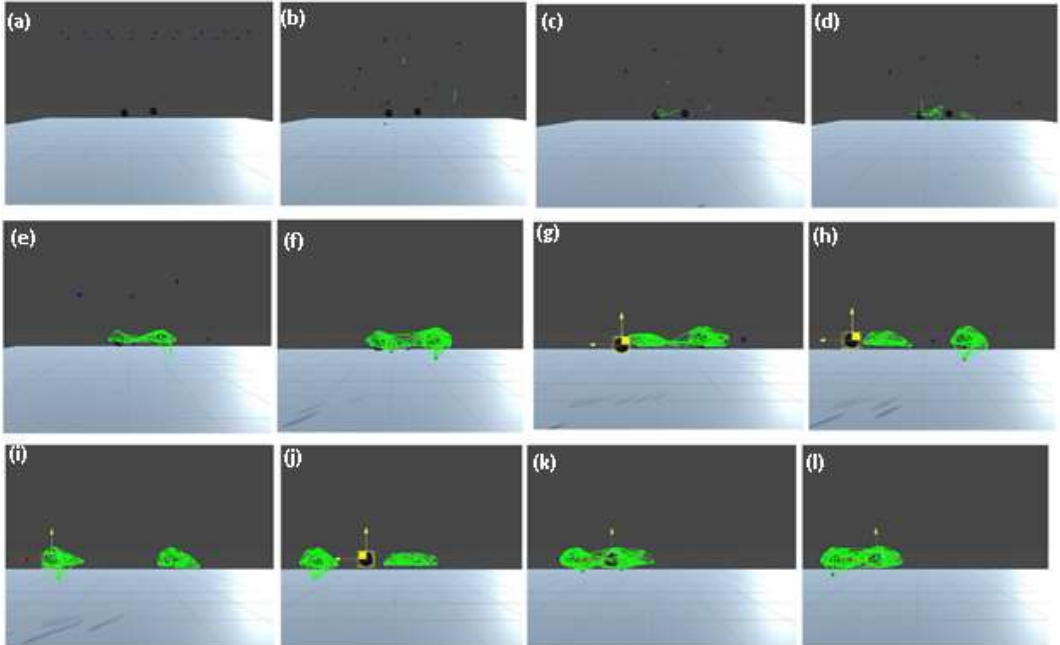

12: Multiple Targets

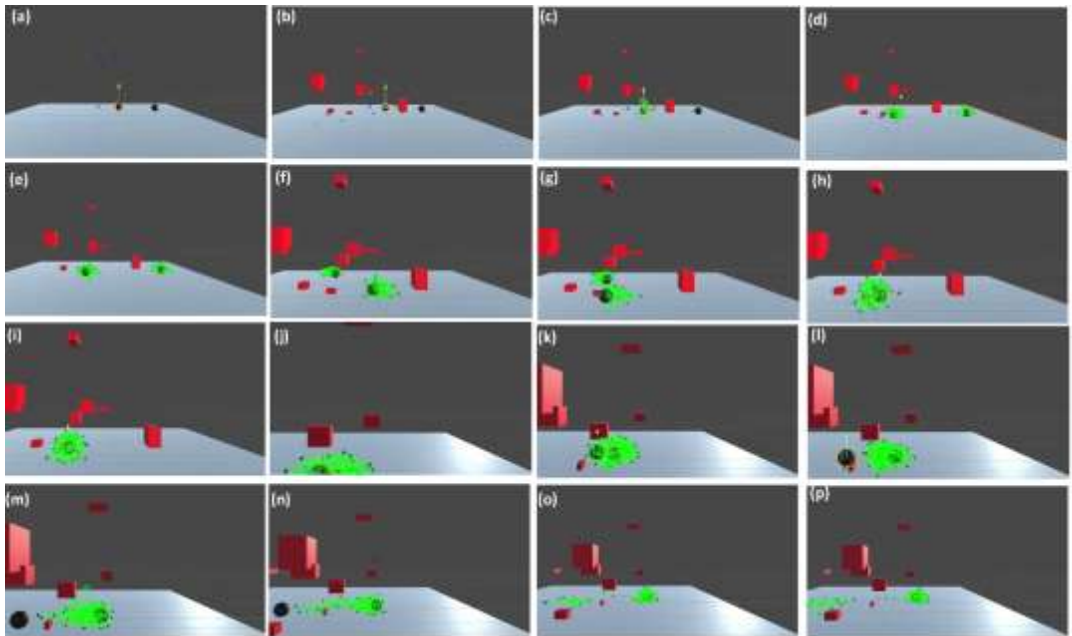

13: Multiple Targets in Complex Environment

To visualize the concepts of large single swarm (Huge Swarm composed of many sub-swarms), dividing that into small swarms and vice versa, distance between the two targets was increased (Fig. 12 (g)). Swarm was able to divide itself into two smaller swarms (Fig. 12 (h-i)). To recreate a single swarm by combining these sub-swarms, targets were brought close together (Fig. 12 (j)). Sub-swarms tailing targets got close to each other and as a result a single swarm was formed (Fig. 12 (k-l)). As shown in Fig. 13, two dynamic targets were added to the environment and, after the swarm localized them, they were made to move away from each other. The swarm divided itself into smaller swarms to track both targets going in different directions. Then, both targets were moved closer to each other. This caused the sub-swarms to gradually merge again to form a single swarm.

### F. EXPERIMENT 6: MULTIPLE TARGETS IN A COMPLEX ENVIRONMENT WITH DYNAMIC OBSTACLES AND ENVIRONMENTAL FACTORS

In this experiment, to check our model's robustness against environmental factors. Factors like gravity, linear drag, and angular drag were modelled in this study. Gravity was modelled at a value of $9.81 m/s^2$ to mimic the earth's surface environment. Linear and angular drag values were set at 0.25N and 0.15Nm, respectively. Despite the addition of these factors, our model navigated the environment successfully with minimal difference compared to its absence as shown in Fig 14.

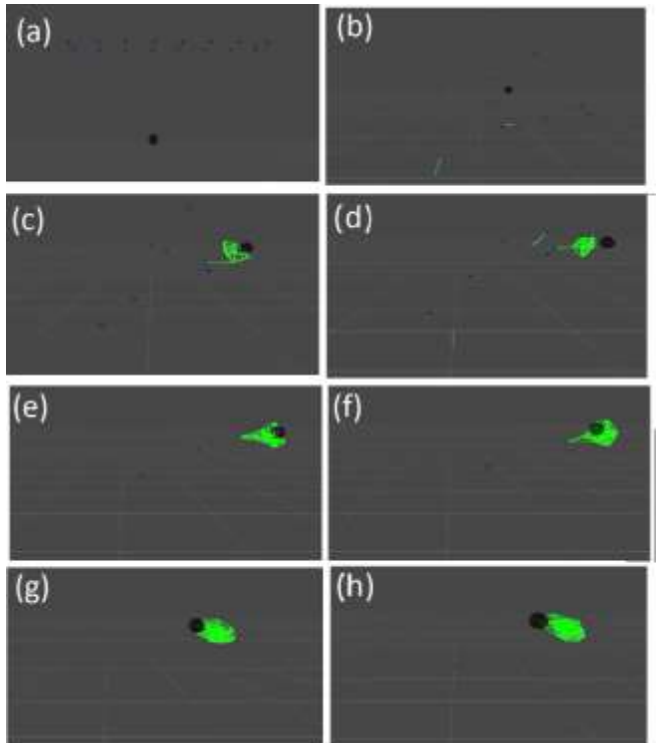

14: Simulating Environmental Factors

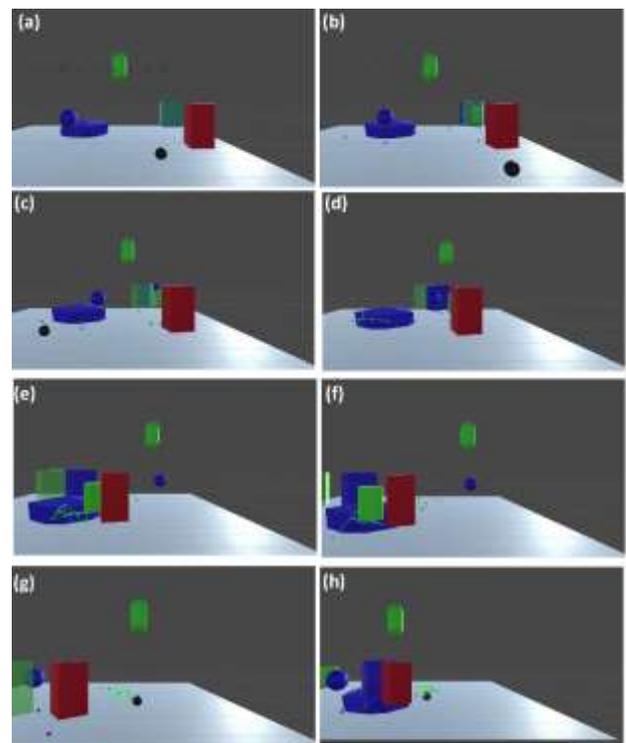

15: Dynamic Obstacle with Fast Moving Target

Dynamic obstacles were introduced as shown in Fig. 15(a). Ground layer here was only present to make visualization better and doesn't have colliders. Target velocity was set as very high and even obstacle movement velocity was identical to the velocity of drone swarm in order to observe behavior in chaotic environments. Drones still exhibited swarm-like formation and behavior (Fig. 15(f-g)), tried to track target but didn't have a lot of success due to high velocity of target. Interestingly drones avoided collisions most of the time using backward movement (Fig. 15(e)) and even moving away from incoming obstacles when necessary (Fig. 15(f)).

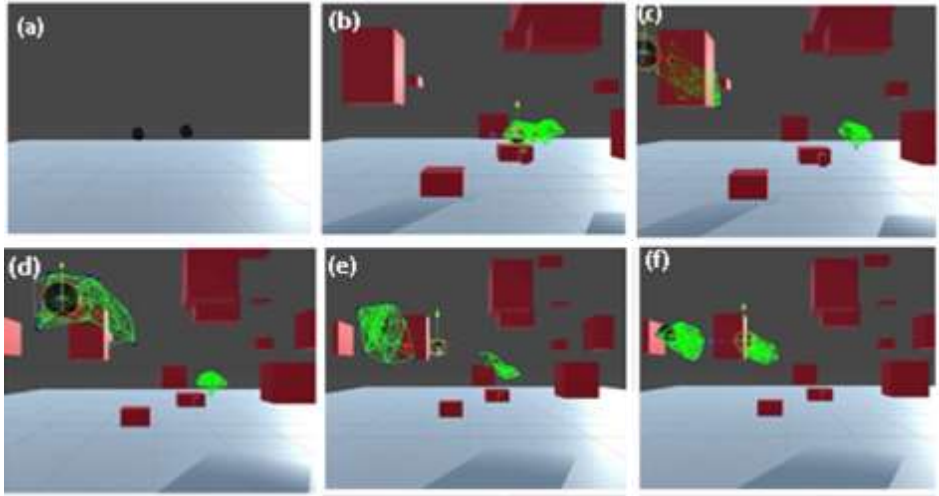

16: Complex Environment

Furthermore, the number of obstacles and targets is increased, object morphology is changed, and some additional environmental factors are added to validate the scalability of the proposed approach, as seen in Fig. 16. Gravitational value is kept at $9.81 m/s^2$, but the linear and angular drag force is increased to a value of 0.37N and 0.25Nm, respectively. Swarm slowed down at drag values exceeding 0.5. Our model was robust enough to tackle this complex environment and found targets and tracked them successfully.

Table 7 Experiments Summary

| S.No. | Experiments | Description |
|---|---|---|
| $01-a$ | Swarm behavior and target localization | Forming a Swarm Organization and maintaining it while locating the target object. |
| $01-b$ | Obstacle avoidance | Agents avoid complex obstacles in a 3D environment. |
| 02 | Target surrounded by obstacles | Obstacles surround the target object from five sides. Only one path is available for drone agents to reach the target object. |
| 03 | Navigation through a small hole | Drone agents passing through a hole single-file. |
| 04 | Target object behind an obstacle | The target object is behind a wall and drone agents must find the shortest path to reach it. |
| $05-a$ | Multiple target objects | Multiple target objects are present, drone swarm needs to surround all of them. |
| $05-b$ | Swarm Subdivision and Combination | Swarm divides itself into smaller swarms known as sub-swarms in other to track multiple targets going in different directions and sub-swarms can combine again when they come close to each other. |
| $06-a$ | Multiple Dynamic Target objects | Targets are all moving, drone swarm not only needs to surround but must keep surrounding it while they are on the move. |
| $06-b$ | Environmental Factors | Drone agents' robustness to environmental factors like gravity, linear drag, angular drag |
| $06-c$ | Multiple target objects in a complex environment | Dynamic multiple target tracking but in a complex environment with a lot of obstacles and environmental factors added in. The number of obstacles is increased along with target objects to check the scalability of our proposed methodology. |

## VI. COMPARATIVE ANALYSIS

Comparison between PPO, TPPO and SAC original architectures and with our customized architectures is presented in Table 8. Number of layers and neurons in each layer are increased to deal with complexity of the training environment while other hyper-parameters are optimized for training efficiency. Furthermore, LSTM is leveraged to help the swarm remember best paths and to aid in obstacle avoidance. Also, comparative results shown in Fig 17 show that our technique performs better than existing techniques.

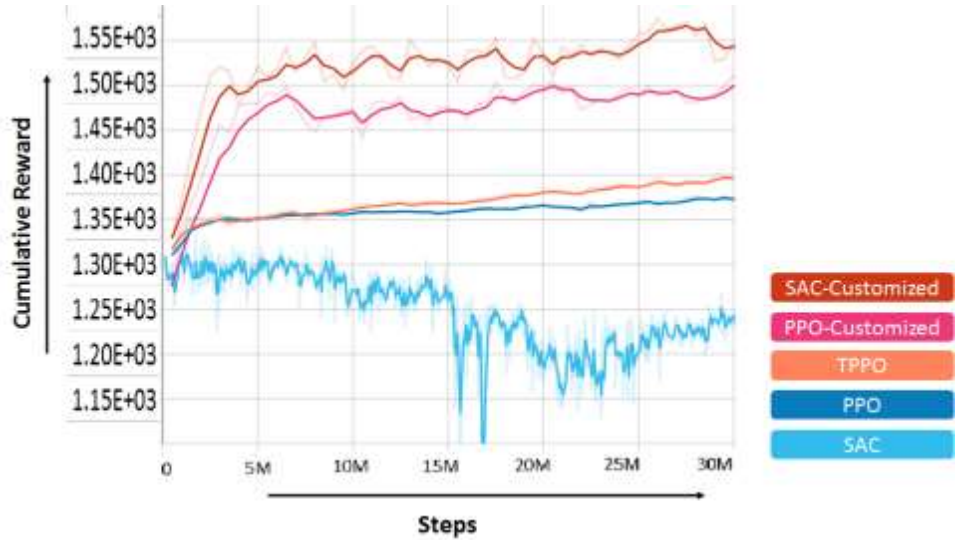

17: Comparison between existing state-of-the-art and our customized models

Table 8 Comparing Algorithms

| Algorithm | Mean Cumulative Reward (CMR) | No. of Layers | Neurons |
|---|---|---|---|
| PPO | 1348 | 2 | 64 |
| TPPO | 1327 | 2 | 128 |
| Customized-PPO | 1448 | 3 | 128 |
| SAC | 1223 | 2 | 256 |
| Customized-SAC | 1550 | 2 | 128 |

VII. CONCLUSION

An efficient methodology to train homogeneous swarm agents is presented for obstacle avoidance and navigation towards multiple targets in complex dynamic 3D environments. A compact vector representation is proposed for presenting state data to our network. This generalizes the behaviors of our drones independent of their proximity to other drones. Furthermore, an appropriate incentive mechanism employing reward functions was developed to design collision-free navigation while maintaining the swarm's connection carefully. Also, the problem of multi-target tracking, where our swarm can track multiple targets while maintaining formation and communication within the swarm was tackled. The concept of dynamic swarms was introduced, where a swarm can be divided to track more than one target simultaneously and also if targets are removed, sub-swarms can combine to form a single larger swarm. Also even when there are multiple targets in the environment in close proximity of each other, sub-swarms can combine into a single swarm for that duration and when targets move away, swarm can again sub-divide. The results demonstrate the approach's universality by putting it to test in various situations that a swarm may face. This strategy can be scaled up to be employed for real-world swarm applications. It can find multiple uses, for example, in search and rescue, food delivery etc. It can also be used against terrorists, for instance tracking their vehicles, autonomous bombardment of their bases. Similarly, it finds both constructive uses such as food and medicine delivery etc., and destructive uses in war, for example, aerial poison bombing etc. It can also be used as a hovering swarm to either protect an entity or to hinder it from making any movements.